\documentclass[runningheads]{llncs}
\usepackage{times}
\usepackage{graphicx}
\usepackage{latexsym}
\usepackage{titletoc}
\usepackage{times}  
\usepackage{helvet} 
\usepackage{courier}  
\usepackage[hyphens]{url}  
\usepackage[misc]{ifsym}
\usepackage{graphicx}  
\usepackage{cite}
\usepackage{amsmath,amssymb,amsfonts}
\usepackage{microtype}
\usepackage{algorithmic}
\usepackage{subfig}
\usepackage{textcomp}
\usepackage{fancyvrb}
\usepackage{xcolor}
\usepackage{multirow}
\usepackage{subfig}
\usepackage[normalem]{ulem}

\usepackage{graphicx}
\begin{document}
\title{Artificial Intelligence for IT Operations (AIOPS) Workshop White Paper}
%
\author{Jasmin Bogatinovski\inst{1}\and
Sasho Nedelkoski\inst{1}\and
Alexander Acker\inst{1}\and
Florian Schmidt\inst{1}\and
Thorsten Wittkopp\inst{1}\and
Soeren Becker\inst{1}\and
Jorge Cardoso\inst{2, 3}\and
Odej Kao\inst{1}
}
%
%
\institute{Distributed and Operating Systems Group, Technische Universität Berlin ,Berlin, Germany \\
\email{firstname.lastname@tu-berlin.de} \and
Huawei Munich Research Center, Munich, Germany \and
University of Coimbra, CISUC, DEI\\
\email{jorge.cardoso@huawei.com}}

\maketitle

\section{Introduction and Motivation}
Large-scale computer systems (e.g., the cloud, IoT and embedded environments) are a key technology that transforms numerous industries including healthcare, finance, manufacturing, education, and transportation.
Billions of devices and users communicate, compute, and store information, and thus depend on the reliability and availability of these systems.
The large computer systems are increasingly adopting new paradigms towards distributed computing that provide higher flexibility of the systems, but also largely increase their complexity. 

Owing to the complexity and inevitable weaknesses in the software and hardware, the systems are prone to failures.
Several studies showed that such failures lead to a decreased reliability, high financial costs, and can impact critical applications~\cite{zhang2015rapid,yuan2014simple,crameri2007staged}. Therefore, loss of control is not allowed for any system or infrastructure, as the quality of service (QoS) is of high importance. 

The large service providers are aware of the need for always-on services with a high availability, and thus already deployed numerous measures such as site reliability engineers and DevOps.
However, the scale and complexity of the computer systems steadily increase to a level where the manual operation becomes infeasible. Operators start using artificial intelligence tools for automation in various operation tasks including system monitoring, anomaly detection, root cause analysis, and recovery.

Artificial Intelligence for IT Operations (AIOps) is an emerging interdisciplinary field arising in the intersection between the research areas of machine learning, big data, streaming analytics, and the management of IT operations. AIOps, as a field, is a candidate to produce the future standard for IT operation management. To that end, AIOps has several challenges. First, it needs to combine separate research branches from other research fields like software reliability engineering. Second, novel modelling techniques are needed to understand the dynamics of different systems. Furthermore, it requires to lay out the basis for assessing: time horizons and uncertainty for imminent SLA violations, the early detection of emerging problems, autonomous remediation, decision making, support of various optimization objectives. Moreover, a good understanding and interpretability of these aiding models are important for building trust between the employed tools and the domain experts. Finally, all this will result in faster adoption of AIOps, further increase the interest in this research field and contribute to bridging the gap towards fully-autonomous operating IT systems. 

The main aim of the AIOPS workshop is to bring together researchers from both academia and industry to present their experiences, results, and work in progress in this field. The workshop aims to strengthen the community and unite it towards the goal of joining the efforts for solving the main challenges the field is currently facing. A consensus and adoption of the principles of openness and reproducibility will boost the research in this emerging area significantly. 

\section{The Landscape of AIOPS}
\begin{figure}[b!]
    \centering
    \resizebox{\textwidth}{!}{\includegraphics{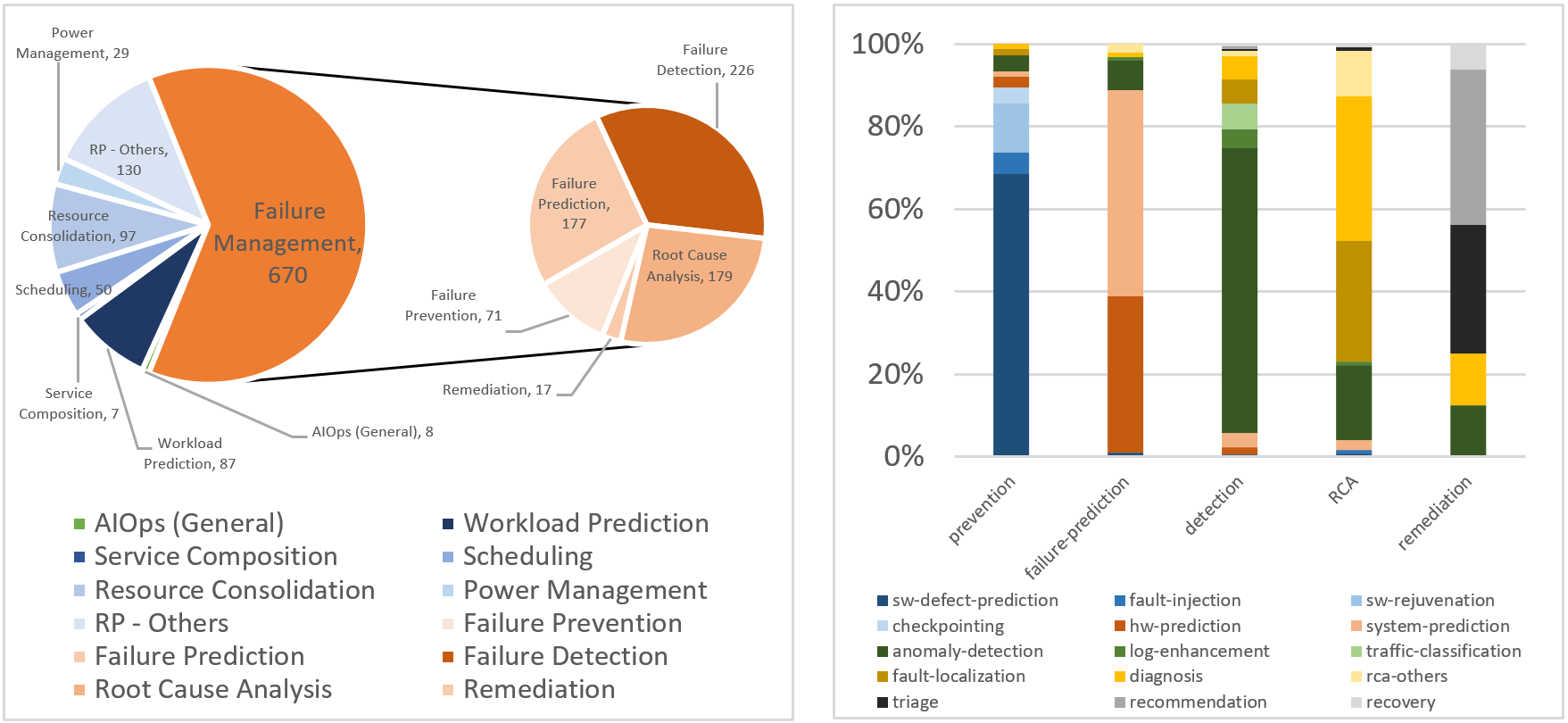}}
    \caption{Left: distribution of AIOps papers in macro-areas and categories. Right: percent distribution of failure management papers by category in corresponding sub-categories.}
    \label{fig:mapping-study_stats}
\end{figure}

\begin{figure}[t!]
 \hspace{-0.29cm}
    \resizebox{1.043\textwidth}{!}{\includegraphics{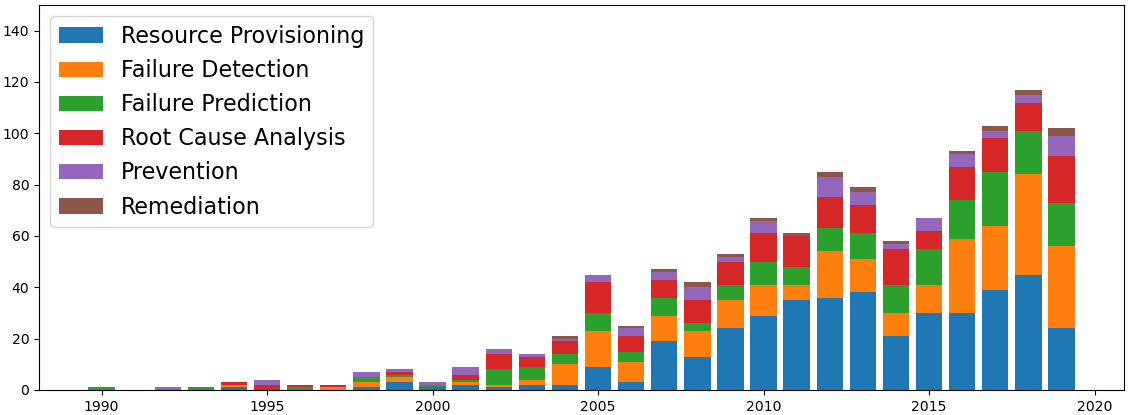}}
    \caption{Published papers in AIOps by year and categories from the described taxonomy.}
    \label{fig:year_category_bar}
\end{figure}

AIOps offers a wide, diverse set of tools for several applications, from efficient resource management and scheduling to complex failure
management tasks such as failure prediction, anomaly detection and remediation. However, being a recent and cross-disciplinary field, AIOps is still
a largely unstructured research area. The existing contributions are scattered
across different conferences and apply different terminology conventions. Moreover, the high number of application areas renders the search and collection of
relevant papers difficult. In the workshop, Notaro et al.~\cite{notaro2020systematic} presented an in-depth analysis of the work concerned with the field of AIOps to cover for these limitations. In this work the authors have identified and extracted over 1000 AIOps contributions through a systematic mapping study, enabling to delineate common trends, problems and tools. \footnote{\figurename~\ref{fig:mapping-study_stats}-\ref{fig:year_category_bar}
, and part of the text in Section 2 was copied from the work of Paolo Notoro et. al. A Systematic Mapping Study in AIOps, In Proceedings Ineternational Conference on Service Oriented Computing, International Workshop on Artificial Intelligence for IT operations 2020, virtual.}


The study highlights the distribution of papers in a presented taxonomy. The left side of Figure \ref{fig:mapping-study_stats} visualizes the distribution of identified papers by macro-area and category. Notaro et al. observe that more than the majority of items (670, 62.1\%) are associated with failure management (FM), with most contributions concentrated in online failure prediction (26.4\%), failure detection (33.7\%), and root cause analysis (26.7\%); the remaining resource provisioning papers support in large part resource consolidation, scheduling and workload prediction. On the right side, it can be observed that the most common problems are software defect prediction, system failure prediction, anomaly detection, fault localization and root cause diagnosis.
To analyze temporal trends present inside the AIOps field, the study measured the number of publications in each category by year of publication. The corresponding bar plot is depicted in Figure \ref{fig:year_category_bar}. Overall, it can be observed that there is increasing interest and growing number of publications in AIOps. Failure detection has gained particular traction in recent years (71 publications for the 2018-2019 period) with a contribution size larger than the entire resource provisioning macro-area (69 publications in the same time frame). Failure detection is followed by root cause analysis (39) and online failure prediction (34), while failure prevention and remediation are the areas with the smallest number of attested contributions (11 and 5, respectively).

\label{sec:conc}

\section{Summary of the studies}
This section consists of three parts. In the first part we briefly describe the 6 papers on the topic of anomaly detection from system data. In the second part we describe the papers on the topic of fault localization and root cause analysis. Finally, in the third part we present novel research directions in AIOps. 

\subsection{Anomaly Detection}
Anomaly detection from system data is one of the most common task~\cite{nedelkoski2020selfattentive,yeanomalycloud, doelitzscheranomalyclouds,nedelkoski2020selfsupervised,du2017deeplog,zhang2019robust}. The main goal is to find the set of interesting observation that is very likely not to be part of the standard behaviour of the system. If these points further lead to some failure in the system they are referred to as anomalies. On this workshop, there were 6 accepted papers on anomaly detection from system data. They included anomaly detection from both numerical and textual data. The considered methods vary with more traditional machine learning techniques with a proclivity towards deep learning technique approaches.

\subsubsection{TELESTO: A Graph Neural Network Model for Anomaly Classification in Cloud Services} Scheinert and Acker~\cite{scheinert2020telesto} propose a novel graph convolutional neural network (GCNN) architecture TELESTO for detecting and classifying anomalies for resource monitoring data streams within large IT-systems. TELESTO is evaluated on a data set generated by injecting synthetic anomalies in a cloud testbed running different IT-services. The evaluation results show that TELESTO outperforms two alternative GCNN architectures.

\subsubsection{Towards Runtime Verification via Event Stream Processing in Cloud Computing Infrastructures} Cotroneo et al.~\cite{cotroneo2020runtime} present a method for runtime verification of cloud computing infrastructures by analysing event streams. Mining is applied to create a set of general rules that represent normal system execution. These are used to verify the system and detect potential anomalies. A preliminary evaluation was performed on OpenStack.

\subsubsection{Online Memory Leak Detection in the Cloud-based Infrastructures}
Jindal et al.~\cite{anshul2020} present the Precog algorithm which is employed for online memory leak detection in cloud-based infrastructures. It observes memory resource allocation over time and compares patterns with a knowledge base of known normal patterns.
Significant deviations are labeled as anomalies. Evaluation on synthetic memory leak data is showing promising results.

\subsubsection{Anomaly Detection at Scale: The Case for Deep Distributional Time Series Models}
Ayed et al.~\cite{ayed2020anomaly} introduce a new methodology for detecting anomalies in time-series data, with a primary application to monitoring the health of (micro-) services and cloud resources. Instead of modelling time series consisting of real values or vectors of real values, the study proposes to model time series of probability distributions. This extension allows the technique to be applied to the common scenario where the data is generated by requests coming into a service, which is then aggregated at a fixed temporal frequency. The results show superior accuracy of the method on synthetic and public real-world data.

\subsubsection{SLMAD: Statistical Learning Based Metric Anomaly Detection}
Shahid et al.~\cite{shahid2020} present a time series anomaly detection framework called Statistical Learning-Based Metric Anomaly Detection (SLMAD) that allows for the detection of anomalies from key performance indicators (KPIs) in streaming time-series data. The method consists of a three-stage pipeline including analysis of time series, dynamic grouping, and model training and evaluation. The experimental results show that the SLMAD can accurately detect anomalies on several benchmark data sets and Huawei production data while maintaining efficient use of resources.

\subsubsection{Using Language Models to Pre-train Features for Optimizing Information Technology Operations Management Tasks} Liu et al.~\cite{liu2020} propose the usage of language models to pre-train relevant features from domain data occurring from IT-operation tasks. The paper presents a case study of detecting anomalies within log-data and demonstrates that domain-specific language models outperform trained models with general-purpose data.

\subsubsection{Multi-Source Anomaly Detection in Distributed IT Systems}~\cite{bogatinovski2021multisource}. The authors utilize
joint representation from the distributed traces and system log data for the task of anomaly detection in distributed systems. The study formalizes a learning task - next template prediction, that is used as a generalization for anomaly detection for both logs and distributed traces. They demonstrate that the joint utilization of traces and logs produced better results compared to the single modality anomaly detection methods. 

\subsection{Fault Localization}
After detecting that there is a fault, an equally important task is to use the data-driven methods to narrow down the set of possible faults that may arise. This task is known as fault localization~\cite{Yuan2019AnAT,zhou2019latent,chen2019empirical,liu2020unsupervised,bogatinovski2020self}. The localization can be done on various data types. On this workshop, approaches from fault localization from metrics and logs were presented. 

\subsubsection{An Influence-based Approach for Root Cause Alarm Discovery in Telecom Networks} Given a set of alarms, Zhang et al.~\cite{zhang2020influence} utilize causality graphs to identify the root cause alarm.  A combination of Hawkes processes, conditional independence testing and graph node embedding is used to construct the causality graphs. An influence ranking method outputs a list of potential root causes. The method outperforms four baselines on synthetic and real-world data.

\subsubsection{Localization of operational faults in cloud applications by mining causal dependencies in logs using Golden Signals}
Agrawal et al.~\cite{aggrawal2020} present an approach for fault detection in cloud applications. Therefore the assumption is, that errors are indicated by the golden signals: Latency, Error, Traffic, and Saturation. First, they mine a causal dependency graph of all nodes which are infected by these golden signals. By utilizing the graph and golden signal errors, they generate a ranked list of possible faulty components. The experiment has shown that faulty components can be detected with quite reliable.

\subsubsection{Discovering Alarm Correlation Rules for Network Fault Management}
Fournier-Viger et al.~\cite{Fournier2020discovering} developed an approach to extract alarm correlation rules in networks. Their approach includes a dynamic attributed graph which reflects the network topology and the included alarms. Thus, they first investigate how alarms are propagated through the network. Then, compression of the number of alarms is performed to figure out root alarms. By revealing the root alarms the number of alarms is decreased effectively addressing the problem of alarm fatigue. 

\subsubsection{Performance Diagnosis in Cloud Microservices using Deep Learning}
Wu et al.~\cite{wu2020performance} introduce an approach to locate the culprits of microservice performance degradation, by investigating the cloud infrastructure on application level and machine level. They mine a service dependency graph which includes the machine and service level of the cloud infrastructure. The anomaly detection is done on metric level. By investigation abnormal metrics and utilizing the service dependency graph the authors are able to detect the root causes for performance degradation in cloud microservices with a high precision of 0.92.

\subsection{Other topics}
Despite, the more traditional tasks of anomaly detection and fault localization on the workshop some papers raise other important issues, like sharing the data between the different IT-services or sharing resources. They can open novel perspectives for the potential of the field. 
\subsubsection{Decentralized Federated Learning Preserves Model and Data Privacy}
Wittkopp and Acker~\cite{wittkopp2020federated} propose a decentralized federated learning approach for sharing knowledge between different IT-services in a privacy-aware procedure. The evaluation shows improvements for log-data anomaly detection when training DeepLog models with a teacher-student approach without sharing directly training data or any model parameters.

\subsubsection{Resource Sharing in Public Cloud System with Evolutionary Multi-agent Artificial Swarm Intelligence}
Chen et al.~\cite{Chen2020resource} present Artificial Swarm Intelligence approach for resource sharing between cloud costumers to minimize resource utilization while guaranteeing Quality of Experience (QoE) to the costumers. The evaluation results show the effectiveness of the approach in optimizing cloud resource allocation for balancing resources to cope with potential peak-time usages.

\section{Conclusion and Future directions}
We observe an increasing interest in the field of AIOps as witnessed by the great number of on-growing publications in AIOps. Some of the tasks, such as failure detection and failure localization show to be predominantly present. However, the large analysis of many studies shows that also there is a lot of work and interest in failure prediction. An important question to be addressed in future is good AIOPS benchmarks. While there exist sets of data for some of the tasks, there is not a public medium where all the data and approaches can be compared. As such, there is a challenge for cross-comparison of methods and approaches in various settings. This prevents the community to keep track of its progress and direction. Following the success stories of many other communities, having such a benchmark is expected to lift the quality of the research and bring all the research groups under one hood. Although the requirements for such a set of benchmark data is not clear due to the specifics of the problem, further effort should be invested in making a public benchmark for AIOps. Furthermore, research and development towards fully autonomous IT operations are required especially in the fields of root cause analysis and self-healing. From the current state of research, there is a gap of solutions that perform well on these important tasks.

\bibliographystyle{splncs04}
\bibliography{main}
\end{document}